\documentclass[twoside]{article}

\usepackage[accepted]{aistats2023}
\usepackage{amsmath}
\usepackage{hyperref}       
\usepackage[capitalise,noabbrev]{cleveref}
\usepackage{url}            
\usepackage{booktabs}       
\usepackage{amsfonts}       
\usepackage{nicefrac}       
\usepackage{microtype}      
\usepackage[usenames, dvipsnames]{xcolor}         
\usepackage{amsthm}         
\usepackage{xspace}
\usepackage{natbib}
\usepackage{bbm}
\usepackage{caption}
\usepackage{subcaption}
\usepackage{graphicx}
\usepackage{enumitem}
\usepackage{dsfont}
\graphicspath{ {./figures/} }

\newcommand{\abs}[1]{\left\lvert#1\right\rvert}

\newcommand{\rvars}[1]{\ensuremath{\mathbf{#1}}\xspace}
\newcommand{\X}{\rvars{X}}

\newcommand{\R}{\mathbb{R}}

\newcommand{\jstate}[1]{\ensuremath{\mathbf{#1}}\xspace}
\newcommand{\e}{\jstate{e}}
\newcommand{\x}{\jstate{x}}
\newcommand{\y}{\jstate{y}}

\newcommand{\val}{\ensuremath{\mathsf{val}}}

\newcommand{\comment}[1]{%
  \text{\phantom{(#1)}} \tag{#1}
}
\newcommand{\eval}{\bigr\rvert}

\theoremstyle{definition}
\newtheorem{thm}{Theorem}

\newtheorem{defn}{Definition}
\newtheorem{lem}{Lemma}
\newtheorem{cor}{Corollary}

\begin{document}

%

%

\twocolumn[

\aistatstitle{Mixtures of All Trees}

\aistatsauthor{ Nikil Roashan Selvam \And Honghua Zhang \And  Guy Van den Broeck }

\aistatsaddress{ 
     UCLA Computer Science \\ \texttt{nikilrselvam@ucla.edu} 
\And UCLA Computer Science \\ \texttt{hzhang19@cs.ucla.edu} 
\And UCLA Computer Science \\ \texttt{guyvdb@cs.ucla.edu} 
}

]

\setcitestyle{authoryear,round,citesep={;},aysep={,},yysep={;}}

\begin{abstract}
Tree-shaped graphical models are widely used for their tractability. However, they unfortunately lack expressive power as they require committing to a particular sparse dependency structure. We propose a novel class of generative models called mixtures of \emph{all} trees: that is, a mixture over all possible~($n^{n-2}$) tree-shaped graphical models over $n$ variables. We show that it is possible to parameterize this Mixture of All Trees (MoAT) model compactly (using a polynomial-size representation) in a way that allows for tractable likelihood computation and optimization via stochastic gradient descent. Furthermore, by leveraging the tractability of tree-shaped models, we devise fast-converging conditional sampling algorithms for approximate inference, even though our theoretical analysis suggests that exact computation of marginals in the MoAT model is NP-hard. Empirically, MoAT achieves state-of-the-art performance on density estimation benchmarks when compared against powerful probabilistic models including hidden Chow-Liu Trees.
\end{abstract}

\section{INTRODUCTION}
Probabilistic graphical models~(PGMs) have been extensively studied due to their ability to exploit structure in complex high-dimensional distributions and yield compact representations. The underlying graph structure of these models typically dictates the trade-off between expressive power and tractable probabilistic inference.  On one end of the spectrum lie tree-shaped graphical models including Chow-Liu trees~\citep{chow-liu}, where the underlying graph is a spanning tree $T=(V,E)$ on $n$ vertices. Tree distributions allow for efficient sampling and exact inference on a variety of queries such as computing marginals~\citep{pearl1988probabilistic, darwiche2003differential} and are widely used in practice~\citep{zhang2017latent}. However, by committing to a single sparse dependency structure (by choice of spanning tree) their expressive power is limited. On the other end of the spectrum, we have densely connected graphical models such as Markov random fields~(MRFs)~\citep{koller2009probabilistic, rabiner1986introduction}, Bayesian networks~\citep{pearl1988probabilistic}, and factor graphs~\citep{loeliger2004introduction}, which excel at modelling arbitrarily complex dependencies~\citep{mansinghka2016crosscat}, but do so at the cost of efficient computation of marginal probabilities.
This spectrum and the underlying tradeoff extends beyond graphical models to generative models at large. For instance, deep generative models like variational autoencoders~(VAEs)~\citep{maaloe2019biva} are extremely expressive, but do not support tractable inference.

In this work, we propose a novel class of probabilistic models called Mixture of \emph{All} Trees (MoAT): a mixture over all possible~($n^{n-2}$) tree-shaped MRFs over $n$ variables; e.g., MoAT represents a mixture over~$10^{196}$ components when modeling joint distributions on $100$ variables. Despite the large number of mixture components, MoAT can be compactly represented by $O(n^2)$ parameters, which are shared across the tree components.
The MoAT model strikes a new balance between expressive power and tractability: 
(i)~it concurrently models all possible tree-shaped dependency structures, thereby greatly boosting expressive power; (ii)~by leveraging the tractability of the spanning tree distributions and the tree-shaped MRFs, it can not only tractably compute \emph{normalized likelihood} but also efficiently estimate \emph{marginal probabilities} via sampling. In addition, as a fixed-structure model, MoAT circumvents the problem of structure learning, which plagues most probabilistic graphical models. 

This paper is organized as follows. Section 2 defines the MoAT model and shows the tractability of exact (normalized) likelihood computation despite the presence of super-exponentially many mixture components. In Section 3, we discuss the MoAT model's parameterization and learning, and demonstrate state-of-the-art performance on density estimation for discrete tabular data. Next, in Section 4, we discuss the tractability of marginals and MAP inference in MoAT and prove hardness results. Finally, we view MoAT as a latent variable model and devise fast-converging importance sampling algorithms that let us leverage the extensive literature on inference in tree distributions. 

\section{MIXTURES OF ALL TREES}
\label{sec:moat}
\begin{figure*}[ht]
    \centering
    \includegraphics[width=0.98\linewidth]{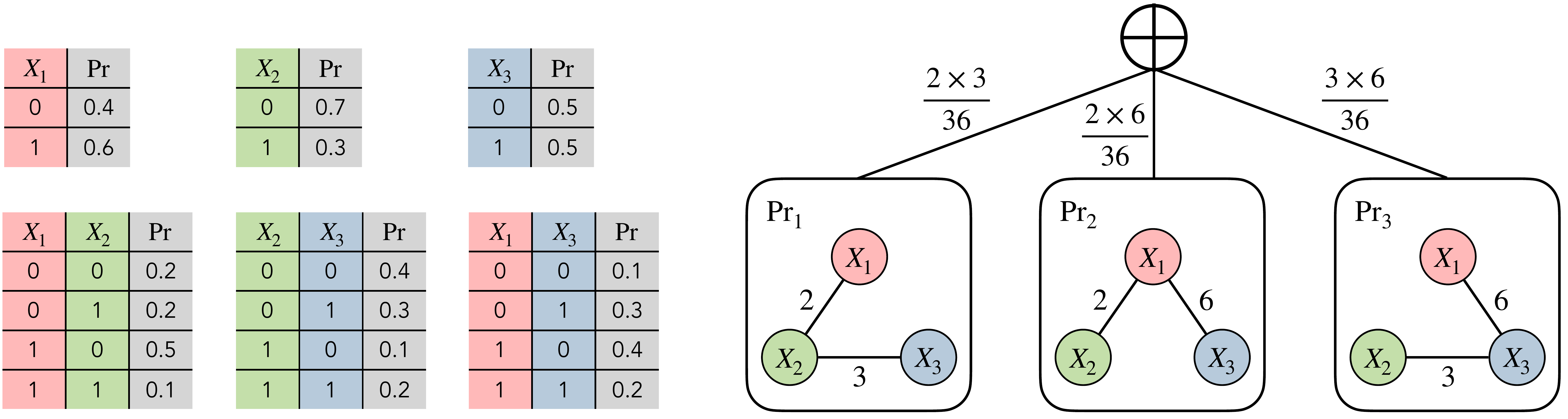}
    \caption{An example MoAT distribution over 3 binary random variables $X_1, X_2, \text{ and }X_3$. The summation at the top denotes a mixture distribution where the mixture weights are given by the weights of the corresponding spanning tree~(shown on the edges). The tables on the left shows the univariate and pairwise marginal distributions, which are shared across the mixture components~(3 possible spanning trees).} 
    \label{fig:moat:example}
\end{figure*}

In this section, we propose mixture of all trees (MoAT) as a new class of probabilistic models. We first introduce tree-shaped Markov random fields~(MRFs) and define the MoAT model as a mixture over all possible tree distributions weighted by the spanning tree distribution. Then, we demonstrate how to tractably compute normalized likelihood on the MoAT model.

\subsection{Mixture of Tree-shaped Graphical Models}
A \emph{tree-shaped MRF} with underlying graph structure $G(V, E)$ represents a joint probability distribution ${\Pr}_{G}$ over $n$ random variables $\mathbf{X} = X_1, \cdots, X_n$ by specifying their univariate and pairwise marginal distributions. Specifically, assuming $G$ is a tree with vertex set $V = \{1, \cdots, n\}$, we associate with each edge $(u, v) \in E$ a pairwise marginal distribution ${P}_{u v}(X_u, X_v)$ and each vertex $u$ a univariate marginal distribution ${P}_{v}(X_v)$. Assuming that ${P}_{uv}$ and ${P}_{v}$ are consistent, then the normalized joint distribution ${\Pr}_{G}$ is given by~\citet{meila-jordan}:
\begin{align}
\label{eq:tree-shape-mrf}
{\Pr}_{G}(\mathbf{x}) = \frac{\prod_{(u, v) \in E}{{P}_{uv}(x_u, x_v)}}{\prod_{u \in V} {P}_{v}(x_v)^{\deg v - 1}},
\end{align}
where $\mathbf{x}\!=\!(x_1, \cdots, x_n)$ denotes assignment to $\mathbf{X}$ and $\deg{v}$ denotes the degree of $v$ in $G$; see $\Pr_{1}(X_1, X_2, X_3)$ in Figure~\ref{fig:moat:example} as an example tree-shaped MRF. 

Despite the tractability of tree-shaped MRFs, they suffer from the problem of limited expressive power. To improve the expressive power, prior works propose to learn mixtures of tree models~\citep{anandkumar2012learning, meila-jordan}, where they focus on simple mixtures of a few trees, and propose EM algorithms for parameter and structure learning. This idea, however, suffers from several limitations. Firstly, while it is known how to optimally pick a single tree distribution with respect to the training data via the Chow-Liu algorithm~\citep{chow-liu}, no known closed form solution exists for picking the optimal set of tree distributions as mixture components from the super-exponentially many possible choices for spanning trees. Secondly, by having a small fixed number of (even possibly optimal) mixture components, the model forces us to commit to a few sparse dependency structures that might not be capable of capturing complex dependencies anyway. 

Though mixture of trees model becomes more expressive as more tree structures are included, the number of parameters increases with the number of mixture components, which seem to suggest that a mixture over a large number of tree components is infeasible. Despite this, we propose the mixture of {\bf all} trees model~(MoAT), a polynomial-size representation for the mixture over all possible (super-exponentially many) tree-shaped MRFs.

Formally, we define:
\begin{align}
\label{eq:moat-definition}
    {\Pr}_{\text{MoAT}}(\x) = \frac{1}{Z} \sum_{T \in \text{ST}(K_n)}  \left(\prod_{e \in T} w_e \right) {\Pr}_{T}(\x)
\end{align}
where $K_n$ denotes the complete graph on $n$ vertices, $\mathsf{ST}(G)$ denotes the set of spanning trees of a connected graph $G$, and $Z$ is the normalization constant. Each mixture component is a tree-shaped MRF $\Pr_{T}$ weighted by $\prod_{e \in T} w_e$, that is, \emph{product of the edge weights of the tree}.
Note that we define the weight of each tree to be proportional to its probability in the \emph{spanning tree distribution}~\citep{borcea2009negative}, which is tractable, allowing for efficient likelihood computation on MoAT~(Section~\ref{subsec:tractable-likelihood}.) 

Though a MoAT model represents a mixture over super-exponentially many tree-shaped MRFs, the number of parameters in MoAT is polynomial-size due the the \emph{parameter sharing} across its mixture components. Specifically, all tree-shaped MRFs $\Pr_{T}(\mathbf{x})$ share the same univariate and pair-wise marginals (i.e., $P_{u}(x_u)$ and $P_{uv}(x_u, x_v)$); in addition, each edge in the graph $K_{n}$ is parameterized by a positive weight $w_{uv}$. To summarize, a MoAT model over $n$ variables has $O(n^{2})$ parameters.

Figure~\ref{fig:moat:example} shows an example MoAT model over 3 binary random variables $X_1, X_2, X_3$, for which there are $3$ possible spanning trees. Note that each of the mixture components (tree distributions) share the same set of marginals, but encode different distributions by virtue of their different dependency structures. 

For example, for the distribution represented in \cref{fig:moat:example},
\begin{align*}
    &{\Pr}_{\text{MoAT}}(X_1=1,X_2=0,X_3=1) \\
    & =\frac{1}{Z} \sum_{T \in \mathsf{ST}(K_n)}  \left(\prod_{e \in T} w_e \right) T(\x) \\
    & =\frac{1}{2 \times 3 + 3 \times 6 + 6 \times 2} \times [ \left(2 \times 3 \times \frac{0.5 \times 0.3}{0.7} \right) \\
    & \quad + \left(2 \times 6 \times \frac{0.5 \times 0.2}{0.6} \right) + \left(3 \times 6 \times \frac{0.2 \times 0.3}{0.5} \right)]
\end{align*}
By Cayley's formula~\citep{chaiken1978matrix}, the number of spanning trees increases super-exponentially with respect to the number of random variables, thus preventing us from evaluating them by enumeration.

\subsection{Tractable Likelihood for MoAT}
\label{subsec:tractable-likelihood}
Despite a super-exponential number ($n^{n-2}$) of mixture components, we show that computing~(normalized) likelihood 
on MoAT is tractable. Our approach primarily leverages the tractability of spanning tree distributions and their compact representation as \emph{probability generating polynomials}, which has been extensively studied in the context of machine learning~\citep{li2016fast, mariet2018exponentiated, robinson2019flexible, ZhangICML21}.
\begin{defn}
Let $\Pr(\cdot)$ be a probability distribution over $n$ binary random variables $\mathbf{X}\!=\!X_1, X_2, \dots,
X_n$, then the \emph{probability generating polynomial} for $\Pr$ is defined as
\begin{align*}
{\sum}_{\mathbf{x}\in \{0,1\}^{n}} \Pr(\mathbf{X} = \mathbf{x}) \left({\prod}_{i \text{ s.t. } x_i = 1} z_i\right),
\end{align*}
where each $z_i$ is an indeterminate associated with $X_i$.
\end{defn}
To define spanning tree distributions and present their representation as probability generating polynomials, we first introduce some notation. Let $G = (V,E)$ be a connected graph with vertex set $V = \{1, \dots, n\}$ and edge set $E$.
Associate to each edge $e \in E$ an indeterminate $z_{e}$ and a weight $w_{e} \in \R_{\geq 0}$.
If $e = \{i, j\}$, let $A_e$ be
the $n \times n$ matrix where $A_{ii} = A_{jj} = 1$, $A_{ij} = A_{ji} = -1$ and all
other entries equal to $0$. Then the \emph{weighted Laplacian} of $G$ is given by
$L(G) = \sum_{e \in E} w_e z_e A_e,$

For instance, the weighted Laplacian for the example MoAT distribution in \cref{fig:moat:example} is
\begin{align*}
    \begin{bmatrix}
    2z_{ab}+6z_{ac} & -2z_{ab} & -6z_{ac}\\
    -2z_{ab} & 2z_{ab}+3z_{bc} & -3z_{bc}\\
    -6z_{ac} & -3z_{bc} & 3z_{bc}+6z_{ac}\\
    \end{bmatrix}
\end{align*}
Using $L(G)_{\backslash\{i\}}$ to denote the principal minor of $L(G)$ that is obtained by removing its $i^{th}$ row and column, by the Matrix Tree Theorem~\citep{chaiken1978matrix}, the probability generating polynomial for the spanning tree distribution is given by:
\begin{align}
\label{eq:st-generating-polynomial}
\operatorname{det}(L(G)_{\backslash\{i\}})=\sum_{T \in \mathsf{ST}(G)} \left(\prod_{e \in T} w_e z_e \right)
\end{align}
Now we derive the formula for computing ${\Pr}_{\text{MoAT}}(\mathbf{x})$ efficiently. We first set $G=K_n$ and $z_e=\frac{P_{uv}}{P_u P_v}$ and define:
$$L^{*} := L(K_n)_{\backslash\{i\}}\eval_{z_e=\frac{P_{uv}}{P_u P_v}};$$
and it follows from Equation~\ref{eq:st-generating-polynomial} that
\begin{align*}
\operatorname{det}\left(L^{*}\right) = \sum_{T \in \mathsf{ST}(K_n)} \left(\prod_{e \in T} w_e \right) \prod_{(u,v) \in T}  \frac{P_{uv}}{P_u P_v};
\end{align*}
note that $\prod_{(u,v) \in T} P_{u} P_{v} = \prod_{u} P_{u}^{\text{deg}(u)}$; hence,
\begin{align*}
&\operatorname{det}\left(L^{*}\right)
\!=\!\frac{1}{\prod_{v \in V} P_v} \sum_{T \in \mathsf{ST}(K_n)}\!\left(\prod_{e \in T} w_e \right)\!\frac{\prod_{(u,v) \in T}P_{uv}}{\prod_{v \in V} P_v^{\operatorname{deg} v-1}}\\
&\quad\!=\!\frac{Z}{\prod_{v \in V} P_v} {\Pr}_{\text{MoAT}},
\end{align*}
where the second equality follows from the definition of MoAT~(Equation~\ref{eq:moat-definition}). Finally, we multiply both sides by $\left(\prod_{v\in V} P_v\right) / Z$ thus ${\Pr}_{\text{MoAT}}(\x)$ can be evaluated as:
\begin{align*}
    {\Pr}_{\text{MoAT}}(\x)
    = \frac{1}{Z} \left(\prod_{v \in V} P_v\left(x_v\right)\right) \operatorname{det}(L^{*} \eval_{\mathbf{x}}).
\end{align*}
Note that the normalization constant of the MoAT model $Z=\sum_{T \in \mathsf{ST}(K_n)}  \left(\prod_{e \in T} w_e \right)$ can be evaluated efficiently as a determinant by replacing the indeterminate $z_e$ with the constant 1. As the computational bottleneck is the determinant calculation, the time complexity is upper bounded as $\mathcal{O}(n^{\omega})$, where $\omega$ is the matrix multiplication exponent.

\section{DENSITY ESTIMATION}
In the previous section, we introduced the MoAT model and described how we can compute likelihood tractably.
In this section, we describe how to parameterize the MoAT model in a way that is amenable to learning and subsequently effective density estimation on real world datasets. There are few desirable properties we seek from this parameterization (of univariate and pairwise marginals in particular). Firstly, we need to parameterize the marginals in way that are \emph{consistent} with each other. This is essential as it guarantees that all tree-shaped mixture components~(Equation~\ref{eq:tree-shape-mrf}) in the MoAT model are normalized. 
Secondly, we want our parameterization to capture the entire space of \emph{consistent} combinations of univariate and pairwise marginals. In particular, this also ensures that every tree distribution is representable by our parameterization.

\subsection{MoAT Parameter Learning}
\begin{figure}[h]
    \centering
    \includegraphics[width=\linewidth]{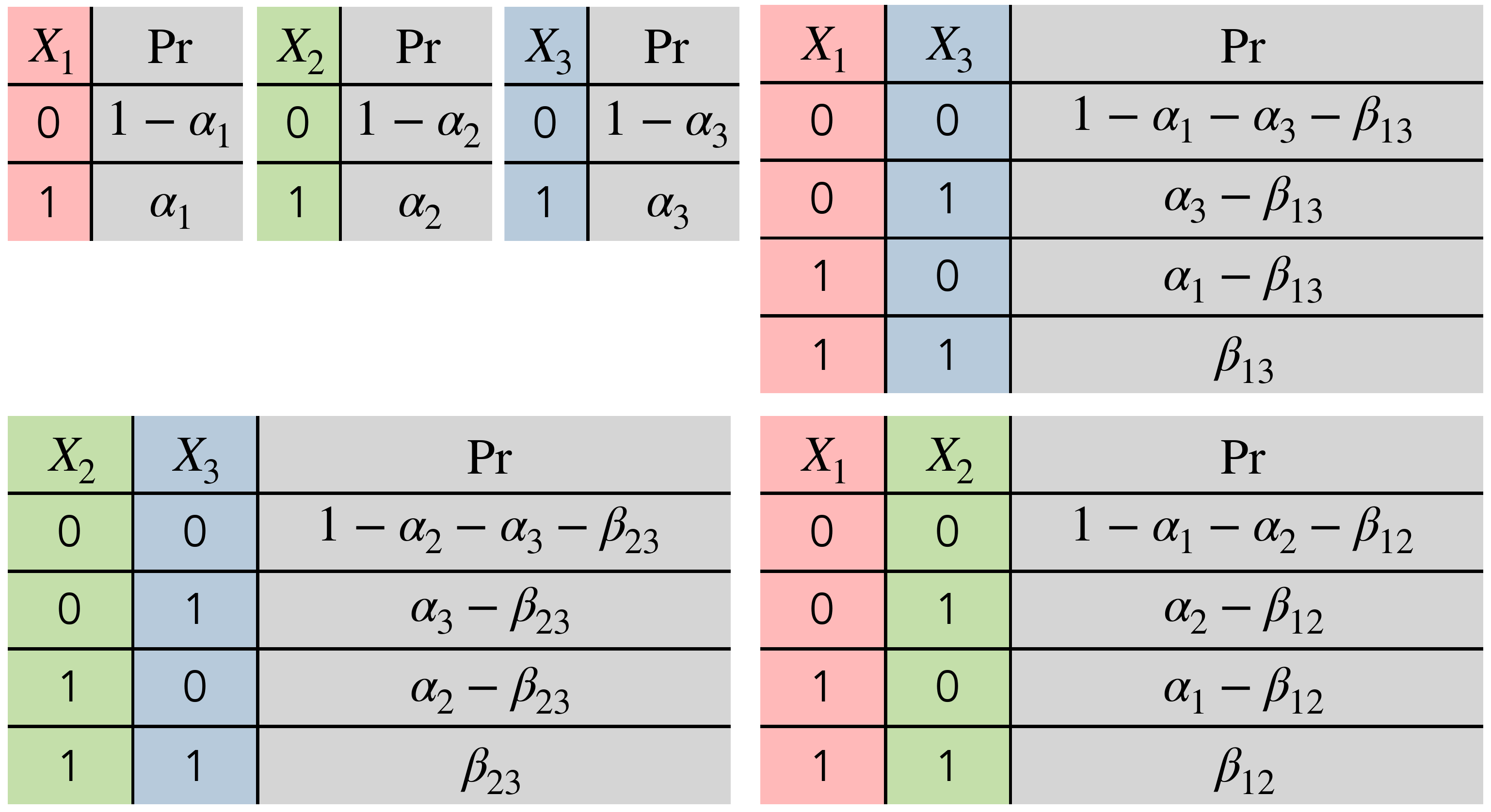}
    \caption{Parameterization for multivariate and univariate marginals for the example distribution on three binary random variables. The $\alpha_{i}$s and $\beta_{ij}$s are the free parameters.} 
    \label{fig:parameterization}
\end{figure}

For a MoAT model over $n$ binary random variables $V=\{X_1, X_2,..., X_n\}$, we propose the following parameterization (as illustrated in \cref{fig:parameterization}):
\begin{itemize}[noitemsep,leftmargin=*]
    \item Edge weights: $w_e \in \mathbb{R}_{\geq 0}$ for $e \in {V \choose 2}$.
    \item Univariate marginals: $p_v\!=\!P(X_v\!=\!1) \in [0,1]$ $\forall v \in V$.
    \item Pairwise marginals: $p_{uv}\!=\!P(X_u\!=\!1, X_v\!=\!1) \in [\max(0,p_u+p_v-1),\min(p_u,p_v)]$ for $\{u,v\} \in {V \choose 2}$.
\end{itemize}

As mentioned in Section~\ref{sec:moat}, to ensure that all the mixture components of MoAT are normalized, our parameterization for $P_{u}$ and $P_{uv}$ needs to be consistent; specifically, they need to satisfy the following constraints:
\begin{itemize}[noitemsep,leftmargin=*]
    \item $P(X_v\!=0)+P(X_v\!=\!1)\!=\!1$ for all $v \in V$.
    \item $\sum_{a\in \{0,1\}}P(X_u\!=\!a,X_v\!=\!b)\!=\!P(X_v\!=b) \text{ }\forall b\in \{0,1\}, \forall \{u,v\} \in {V \choose 2}$.
    \item $\sum_{(a,b)\in \{0,1\}^{2}}P(X_u\!=\!a,X_v\!=\!b)\!=\!1 \text{ }\forall \{u,v\} \in {V \choose 2}$.
\end{itemize}

\begin{lem}
For any distribution $\Pr(\cdot)$ over binary random variables $X_1, \dots, X_n$, there exists a set of parameters~(i.e., $p_v$ and $p_{uv}$) in our hypothesis space such that $\Pr(X_u) = P_{u}(X_u)$ and $\Pr(X_u, X_v) = P_{uv}(X_u, X_v)$ for all $1 \leq u, v \leq n$; i.e., the univariate and pair-wise marginals of $\Pr$ are the same as  $P_{u}$ and and $P_{uv}$.
\end{lem}

See appendix for proof. This lemma shows that the MoAT parameterization is not just valid, but also fully general in the sense that it covers all possible \emph{consistent} combinations of univariate and pairwise marginals.
Further, the MoAT parameterization naturally extends to categorical variables. For categorical random variables $V=\{X_1, X_2,..., X_n\}$, let $\val(X_i)=\{1,2,\cdots , k_i\}$. It is easy to see that the values $P(X_v=i)$ for $i \in \{1,2,\cdots , k_v-1\}$ uniquely determine the univariate marginals. Similarly, the values $P(X_u\!=\!i, X_v\!=\!j)$ for $(i,j) \in \{1,2,\cdots , k_u-1\} \times \{1,2,\cdots , k_v-1\}$ uniquely determine the pairwise marginals. This extension is provably valid, but not fully general.
For MoAT over categorical variables, whether there exists a fully general parameterization~(i.e., Lemma 1 holds) is unknown. See appendix for a detailed discussion.

\paragraph{Parameter Learning} For individual tree distributions, the optimal tree structure (as measured by KL divergence from training data) is the maximum weight spanning tree of the complete graph, where edge weights are given by mutual information between the corresponding pairs of variables~\citep{chow-liu}. Following this intuition, we use mutual information to initialize $w_e$; besides, we also initialize the univariate and pairwise marginals of the MoAT model by estimating them from training data. Finally, given our parameter initialization, we train the MoAT model by performing maximum likelihood estimation~(MLE) via stochastic gradient descent. 

\begin{figure}[h]
\centering
     \begin{subfigure}[b]{0.32\textwidth}
         \centering
         \includegraphics[width=0.95\columnwidth]{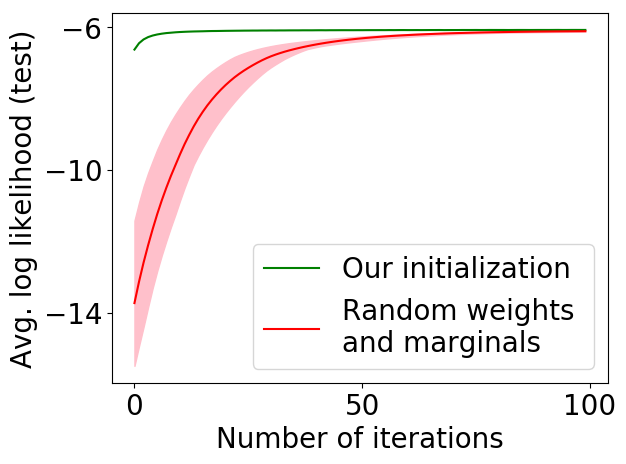}
         \caption{NLTCS Dataset}
         \label{fig:ablation1}
     \end{subfigure}
      \begin{subfigure}[b]{0.32\textwidth}
         \centering
         \includegraphics[width=0.95\columnwidth]{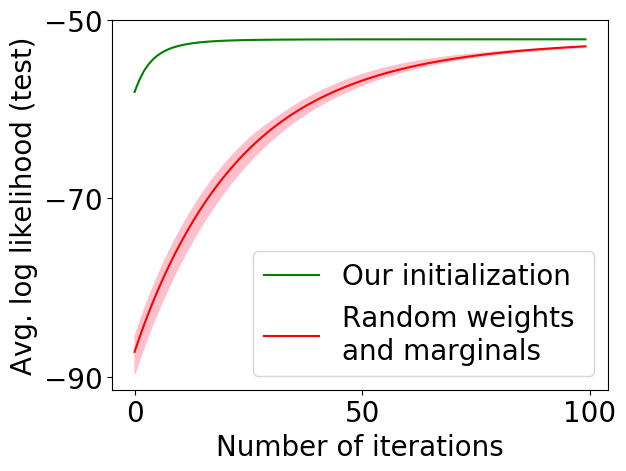}
         \caption{Jester Dataset}
         \label{fig:ablation2}
     \end{subfigure}
{\caption{{Average log-likelihood throughout training on datasets across various data dimensionalities: our initialization vs. random initialization~(averaged over 5 runs).}}
\label{fig:init}}
\end{figure}

It is worth noting that \emph{our parameter initialization is deterministic}. We perform ablation studies to check the effectiveness of our initialization. As shown in Figure~\ref{fig:init}, compared to random initialization, we observe that our
special initialization always leads to better initial log likelihood, faster convergence and better final log likelihood.

\subsection{Density Estimation via MoAT}
\begin{table}[]
\centering
\begin{tabular}{l|l| lll}
\toprule
Dataset   & \# vars & MoAT & HCLT    & MT     \\ \hline 
nltcs     & 16 & -6.07     & \textbf{-5.99}   & -6.01   \\
msnbc     & 17 & -6.43     & \textbf{-6.05}   & -6.07   \\
kdd       & 65 & \textbf{-2.13}     & -2.18   & \textbf{-2.13}   \\
plants    & 69 & -13.50     & -14.26  & \textbf{-12.95}  \\
baudio    & 100 & \textbf{-39.03}     & -39.77  & -40.08  \\
jester    & 100 & \textbf{-51.65}    & -52.46  & -53.08  \\
bnetflix  & 100 & \textbf{-55.52 }    & -56.27  & -56.74  \\
accidents & 111 & -31.59     & \textbf{-26.74}  & -29.63  \\
tretail   & 135 & \textbf{-10.81}     & -10.84  & -10.83 \\
pumsb     & 163 & -29.89     & \textbf{-23.64}  & -23.71   \\
dna       & 180 & -87.10     & \textbf{-79.05}  & -85.14  \\
kosarek   & 190 & \textbf{-10.57}     & -10.66  & -10.62   \\
msweb     & 294 & \textbf{-9.80}     & -9.98   & -9.85   \\
book      & 500 & \textbf{-33.46}     & -33.83  & -34.63  \\
tmovie    & 500 & \textbf{-49.37}     & -50.81  & -54.60  \\
cwebkb    & 839 & \textbf{-147.70}     & -152.77 & -156.86 \\
cr52      & 889 & \textbf{-84.78} & -86.26  & -85.90  \\
c20ng     & 910 & \textbf{-149.44} & -153.4  & -154.24 \\
bbc       & 1058 & \textbf{-243.82} & -251.04 & -261.84 \\
ad        & 1556 & \textbf{-15.30} & -16.07  & -16.02     \\

\bottomrule
\end{tabular}
\caption{Comparison of average log likelihood of MoAT, HCLT, and MT across the Twenty Datasets benchmarks. Best results are presented in bold.}
\label{tab:density}
\end{table}

We evaluate MoAT on a suite of density estimation datasets called the Twenty Datasets ~\citep{twenty-datasets}, which contains 20 real-world datasets covering a wide range of application domains including media, medicine, and retail. This benchmark has been extensively used to evaluate tractable probabilistic models. We compare MoAT against two baselines:~(1)~hidden Chow-Liu trees~(HCLTs)~\citep{hclt}, which are a class of probabilistic models that achieve state-of-the-art performance on the Twenty Datasets benchmark
and~(2)~the mixture of trees model~(MT)~\citep{meila-jordan}.

\cref{tab:density} summarizes the experiment results. MoAT outperforms both HCLT and MT on 14 out of 20 datasets. In particular, the MoAT model beats baselines by large margins on all datasets with more than 180 random variables. It is also worth noting that despite having fewer parameters ($\mathcal{O}(n^2)$) than MT ($\mathcal{O}(k\cdot n^2)$, where $k$ is the number of mixture components in MT), MoAT almost always outperforms MT, with the exception of a few smaller datasets, where MoAT does not have enough parameters to fit the data well.

\section{ON THE HARDNESS OF MARGINALS AND MAP INFERENCE}
In this section, we prove the hardness of semiring queries (which is a generalization of marginals) and  Maximum a posteriori~(MAP) inference on the MoAT model.
\subsection{On the Hardness of Computing Marginals}
\label{sec:marginal-hardness}
First, we define the notion of semiring queries.
\defn{Semiring Queries (SQ): Let $p(\mathbf{X})$ be a real-valued function over random variables $\mathbf{X}$. The class of semiring queries $\mathcal{Q}_{\mathrm{F}}$ is the set of queries that compute values of the following form:
\begin{align*}
f(\boldsymbol{e})=&\sum_\mathbf{z}p(\mathbf{z}, \boldsymbol{e})
\end{align*}
where $e \in \mathrm{val}(\mathbf{E})$ is a partial configuration for any subset of random variables  $\mathbf{E} \subseteq \mathbf{X}$, and $\mathbf{Z}=\mathbf{X} \backslash \mathbf{E}$ is the set of remaining random variables.}

When the semiring sum/product operations correspond to the regular sum/product operations and the function $p$ is a likelihood function, the semiring query $f(\e)$ actually computes marginal probabilities. 

In fact, if $p$ is the likelihood function for the MoAT model, for an assignment $\mathbf{e}$ to $\mathbf{E} \subseteq \mathbf{X}$,
\begin{align*}
f(\boldsymbol{e})=& \frac{1}{Z} \sum_\mathbf{z} \sum_{T \in \mathsf{ST}(K_n)} \left(\prod_{v \in V} P_v\left(x_v\right)\right) \\
& \left(\prod_{(u,v) \in T} w_{(u,v)} \frac{P_{uv}(x_u,x_v)}{P_u(x_u) P_v(x_v)} \right),
\end{align*}
where $Z=\sum_{T \in \mathsf{ST}(K_n)}  \left(\prod_{e \in T} w_e \right)$ is the normalization constant and $\mathbf{z}$ enumerates over all instantiations of $\mathbf{Z}=\mathbf{X} \backslash \mathbf{E}$. Thus, in this case, $f(\e)$ actually computes marginals in the MoAT model. However, the generality of the semiring queries allows for negative parameter values and hence negative ``probabilities'', which we leverage to prove hardness of semiring queries on the MoAT model.

Since most marginal computation algorithms on tractable probabilistic models (such as the jointree algorithm which relies on variable elimination \citep{ZhangPoole, Dechter1996} and circuit compilation based methods \citep{chavira,darwiche2002factor}) are semiring generalizable~\citep{wachter,kimmig17,bacchus09}, the hardness of semiring queries on the MoAT model would strongly suggest the hardness of marginal computation. In other words, the hardness of semiring queries would rule out most marginal inference techniques in the literature as they perform purely algebraic computations on the parameter values without any  restrictions/assumptions on the range of these values. We dedicate the rest of this subsection to establishing the same, while deferring most technical proof details to the appendix.
\thm{Computation of semiring queries on the MoAT model is NP-hard.}
\begin{proof}
To prove the hardness of semiring queries, we proceed by a reduction from the subset spanning tree problem~(denoted $\mathsf{SST}$), which we define below. 
\lem{Define $\mathsf{SST}$ as the following decision problem: given a connected graph $G=(V,E)$ and a subset $K \subset V$ of the vertices, decide if there exists a spanning tree of $G$ whose leaves are exactly $K$. $\mathsf{SST}$ is NP-hard.}

Consider an arbitrary connected graph $G=(V,E)$ on $|V|=n \geq 3$ vertices and a subset of vertices $K \subset V$. Set MoAT likelihood function parameters on $n$ binary random variables $\X=\{X_1,X_2,\ldots, X_n\}$ (corresponding to the vertices of G) as follows:
\begin{itemize}
    \item $ 0 <\epsilon < 1$
    \item $w_e= \begin{cases} 
      1, & e=\{i,j\} \in E \\
      0, & \text{otherwise}
   \end{cases}$
   \item $\mathit{val}(X_i)=\{0,1\}$. 
   \item $P_v(0)=\epsilon, P_v(1)=-1$ for all $v \in V$.
   \item $P_{uv}(\alpha, \beta)=\begin{cases} 
        \epsilon, & \alpha = \beta = 0  \\
        0, & \alpha = \beta = 1 \\
        -\epsilon, &\text{otherwise} \\
   \end{cases}$
\end{itemize}
One can intuitively interpret an assignment of $1$ as corresponding to labelling a node as a leaf, and $0$ as marking it as unknown. The univariate and pairwise marginals have been carefully chosen to ensure that any tree assigns higher probability to assignments where all the nodes assigned $1$ are leaves in the tree and lower probabilities to assignments where one or more nodes that are assigned $1$ are actually internal nodes. In fact, for any tree, there exists a likelihood separation of $\epsilon$ between assignments that agree on the leaves and those that do not. By assigning $1$ to all the variables in $K \subset V$ and $0$ to others, and by choosing a sufficiently small $\epsilon$, we can now effectively use the MoAT likelihood as an indicator for the presence of an spanning tree whose leaves are a superset of $K$. More impressively, we can exactly count the number of spanning trees that satisfy the desired property, and we formalize the same in the following lemma.

\lem{Let $\x$ be a complete assignment, and denote by $\mathsf{ONES}(\x)$ the set of variables are are set to $1$ in $\x$. Denote by $\abs{\x}$ the value $\abs{\mathsf{ONES}(\x)}$ and $\mathsf{LEAVES}(T)$ the set of leaves of a spanning tree $T$. Let $k$ be the number of spanning trees $T$ of $G$ with $\mathsf{ONES}(\x) \subseteq \mathsf{LEAVES} (T)$. Then, 
$$\begin{cases} 
        \frac{k}{\epsilon^{n-2}} \leq Z \cdot p(\x) \leq \frac{k}{\epsilon^{n-2}} +\frac{n^{n-2}}{\epsilon^{n-3}}, &  \abs{\x}\%2=0\\
        \frac{-k}{\epsilon^{n-2}}+\frac{-n^{n-2}}{\epsilon^{n-3}} \leq Z \cdot p(\x) \leq \frac{-k}{\epsilon^{n-2}} , & \abs{\x}\%2=1
        \\
   \end{cases}$$
}
See appendix for proof.
\cor{Let $\epsilon < \frac{1}{2^{n+1}\cdot n^{n-2}}$. The number of spanning trees $T$ with $K \subseteq \mathsf{LEAVES}(T)$ is given by $\abs{\lfloor Z \cdot \epsilon^{n-2} \cdot p(\x) \rceil}$, where $x_i=1$ if and only if $i \in K$ (that is, $\x$ is the assignment that assigns $1$ to all the variables in $K$ and $0$ to all the other variables), $\lfloor x \rceil$ denotes the closest integer to $x$.}
\label{cor:count-trees}
\begin{proof}
    Let $k$ be the number of spanning trees $T$ with $K \subseteq \mathsf{LEAVES}(T)$. When $\abs{\x}\%2=0$, $ k \leq Z \epsilon^{n-2} p(\x) \leq k + \epsilon n^{n-2} \leq k + \frac{1}{2^{n+1}}$. Thus, $\abs{\lfloor Z \epsilon^{n-2} p(\x) \rceil}=k$ as desired. An analogous proof holds for the case of $\abs{\x}\%2=1$.
\end{proof}
Note that $\frac{P_{uv}}{P_u P_v} \geq 0$, and hence the sign of $p(\x)$ depends solely on the parity of $\abs{\x}$. Thus, we can leverage the inclusion-exclusion formula to count spanning trees $T$ with $K = \mathsf{LEAVES}(T)$ using expressions for number of spanning trees $T$ with $K \subseteq \mathsf{LEAVES}(T)$ given by Corollary~\ref{cor:count-trees}. 

\lem{The number of spanning trees $T$ with $K = \mathsf{LEAVES}(T)$ is given by $\abs{\lfloor Z \epsilon^{n-2} f(\e) \rceil}$. } 

\begin{proof}[Proof Sketch]
From the inclusion-exclusion formula we obtain that the number of spanning trees $T$ with $K = \mathsf{LEAVES}(T)$ (upto sign) is given by
\begin{align*}
    & \sum_{K\subseteq L} (-1)^{\abs{L}}\sum_{T \in \mathsf{ST}(G)} \mathds{1}(L \subseteq \mathsf{LEAVES}(T)) \\
    &\quad =\sum_{val(z_1)} \sum_{val(z_2)} \ldots \sum_{val(z_k)} (-1)^{\abs{\x}}\abs{\lfloor Z \epsilon^{n-2} p(\x) \rceil} \\
    &\quad =\lfloor Z \epsilon^{n-2} \sum_{val(z_1)} \sum_{val(z_2)} \ldots \sum_{val(z_k)}   p(\x) \rceil \\ 
    &\quad =\lfloor Z \epsilon^{n-2} f(\e) \rceil
\end{align*}
\end{proof}

We now obtain that there exists a spanning tree $T$ with $K = \mathsf{LEAVES}(T)$ if and only if $\abs{\lfloor Z \epsilon^{n-2} f(\e) \rceil} > 0$. This completes the reduction from $\mathsf{SST}$, as desired.
\end{proof}

It is worth re-emphasizing the strength of this hardness result in the context of marginal computation, in that it eliminates all marginal inference algorithms that are agnostic to parameter values (which is, to the best of our knowledge, all possible known exact marginal inference techniques in literature). This opens up an interesting question about new classes of marginal computation algorithms that are not parameter-value agnostic.

\subsection{On the Hardness of MAP Inference}
In this section, we prove that maximum-a-posteriori~(MAP) inference~(i.e., computing the most likely assignment) for the MoAT model is NP-hard via a reduction from the 3-coloring problem~\citep{lovasz1973coverings}.
\thm{MAP inference for MoAT is NP-hard}.
\begin{proof}
Consider an arbitrary connected graph $G=(V,E)$ on $|V|=n$ vertices. Build a MoAT model $M$ on $n$ discrete random variables $\X=\{X_1,X_2,\ldots, X_n\}$ (corresponding to the vertices of G) as follows:

\begin{itemize}
    \item $w_e= \begin{cases} 
      1, & e=\{i,j\} \in E \\
      0, & \text{otherwise}
   \end{cases}$
   
   \item $\mathit{val}(X_i)=\{R,G,B\}$. 
   \item $P_v(R)=P_v(B)=P_v(G)=\frac{1}{3}$ for all $v \in V$.
   \item $P_{uv}(\alpha, \beta)=\begin{cases} 
        0, & \alpha = \beta \\
        \frac{1}{6}, & \alpha \neq \beta 
   \end{cases}$
   
\end{itemize}

Observe that the weights define a uniform distribution over all possible spaninng trees of $G$. Furthermore, the univariate marginals $P_v$ and pairwise marginals $P_{uv}$ are consistent and define a valid tree distribution.

Next, observe that a complete assignment $\x$ to $\X$ corresponds to a coloring of the original graph G. It is easy to check that for any particular spanning tree T, \\
$T(\x)=\begin{cases} 
        \frac{1}{3\times 2^{n-1}}, & \x \text{ is a valid 3-coloring of the tree} \\
        0, & \text{otherwise}
   \end{cases}$
   
  Now, we show that $\x$ is a valid 3 coloring of the given graph $G$ if and only if $M(\x)=\frac{1}{3\times 2^{n-1}}$.

    Firstly, if $\x$ is a valid 3-coloring of $G$, then no pair of adjacent vertices in $G$ are assigned the same color. Hence, the probability assigned to $\x$ by any of the spanning trees of $G$ is $\frac{1}{3\times 2^{n-1}}$. Hence, 
    \begin{align*}
        M(\x) &= \frac{1}{Z} \sum_{T \in \mathsf{ST}(G)} \left(\prod_{e \in T} w_e \right)   \frac{\prod_{(u,v) \in T}P_{uv}\left(x_u, x_v\right)}{\prod_{v \in V} P_v\left(x_v\right)^{\operatorname{deg} v-1}}\\
        =& \frac{1}{3\times 2^{n-1}} \left( \frac{1}{Z} \sum_{T \in \mathsf{ST}(G)} \left(\prod_{e \in T} w_e \right) \right) 
        = \frac{1}{3\times 2^{n-1}}
    \end{align*}    
    Conversely, if $\x$ is not a valid 3-coloring of $G$, then there exist at least one pair of neighboring vertices in $G$ which share the same color. Now, any spanning tree that contains the corresponding edge (which always exists) would assign zero likelihood to $\x$ and $M(\x)$ be strictly less than $\frac{1}{3\times 2^{n-1}}$.\\
Thus, the graph is 3-colorable if and only if the global MAP state of $M$ has a probability of $\frac{1}{3\times 2^{n-1}}$.
\end{proof}

\section{EFFICIENT APPROXIMATE INFERENCE}

Unlike usual mixture models, all mixture components in MoAT are close to maximum likelihood on the entire dataset (owing to their consistent univariate and pairwise marginals), but are just sufficiently different enough to model complex dependencies. In this section, we explore how this key observation combined with the tractability of tree-shaped models lets us devise fast-converging algorithms for approximate inference on MoAT.

\subsection{MoAT as a Latent Variable Model}
Interestingly, the MoAT model yields itself to being interpreted as a latent variable model in an extremely natural way with clear semantics. Defining $Y$ to be the latent random variable with $\val(Y)=\mathsf{ST}(G)$, one can view MoAT as a distribution over $\{Y,X_1, X_2,..., X_n\}$, where $Y$ models the choice of spanning tree, and inference of the form ${\Pr}_{\text{MoAT}}(\x)$ amounts to marginalizing out the latent variable~$Y$. More precisely, 
\begin{align*}
    {\Pr}_{\text{MoAT}}(\x)
    \!=&\!\sum_{T \in \mathsf{ST}(G)} \frac{\left(\prod_{e \in T} w_e \right)}{Z}  \cdot \frac{\prod_{(u,v) \in T}{\Pr}_{uv}\left(x_u, x_v\right)}{\prod_{v \in V} P_v\left(x_v\right)^{\operatorname{deg} v-1}} \\
    =& \sum_{y \in \val(Y)} \quad P(y) \quad \quad \cdot \qquad P(\x \mid y)
\end{align*}
It is worth emphasizing the distinctiveness of this characterization. Typically in latent variable models, the latent variables act as higher dimensional features over some subset of the variables. However, for the MoAT model, the latent variable controls the sparse dependency structure that is enforced across the same set of variables.

\begin{figure*}[t]
    \centering
    \includegraphics[scale=0.275,trim={0 0.33cm 0 0},clip]{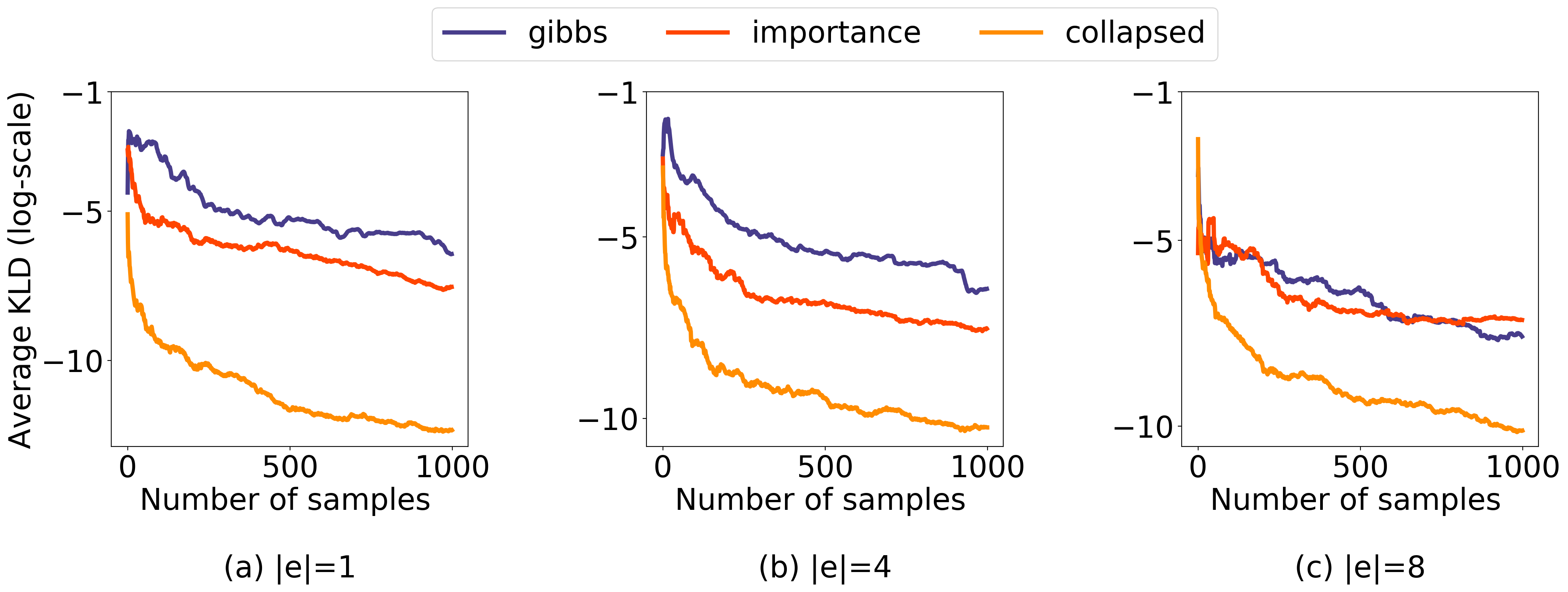}
    \caption{Convergence of various sampling algorithms for posterior marginal inference on NLTCS with different evidence sizes. The reported results are averaged across 5 random seeds. }
    \label{fig:kld-convergence}
\end{figure*}

\subsection{Efficient Importance Sampling on MoAT}
Exact marginals and conditionals are provably tractable on tree distributions owing to classic techniques such as variable elimination. Consequently, tree distributions are extremely amenable to efficient conditional sampling~\citep{koller2009probabilistic}. We show that MoAT, a mixture over tree distributions, also supports effective 
conditional sampling even though our theoretical analysis (\cref{sec:marginal-hardness}) suggests that even computation of marginals in MoAT is NP-hard. 

\paragraph{Importance Sampling} Revisiting the view of MoAT as a latent variable model $P(Y,\X)$, we arrive at a very natural choice of proposal distribution $Q(Y,\X)$ that leads to an efficient importance sampling algorithm~\citep{Tokdar2010ImportanceSA}. For evidence $\e$, (and abusing notation to have $\x$ refer to an assignment to the unobserved variables) we have that:
\begin{align*}
&Q(y,\x \mid \e) \\
&\quad = P(y)P(\x \mid \y\e) \approx P(y\mid \e)P(\x \mid \y\e) = P(y,\x \mid \e)
\end{align*}
At a high level, this amounts to sampling a spanning tree unconditionally~\citep{sample-spanning-tree}, and then sampling the remaining variables from the chosen tree distribution conditioned on the evidence. More precisely, the weighting function for the samples drawn from the proposal distribution is given by
\begin{align*}
w(y,\x \mid \e)
=\frac{P(y,\x \mid \e)}{Q(y,\x \mid \e)} 
=\frac{P(y \mid \e)}{P(y)} = \frac{P(\e \mid y)}{P(\e)}
\end{align*}
The efficiency of the sampling algorithm (as evaluated through, say, the effective sample size) depends on how close the sample weights are to $1$. Intuitively, the ratio $\frac{P(\e \mid y)}{P(\e)}$ captures how much the likelihood of partial evidence in a single spanning tree differs from the corresponding likelihood in the model. As all the mixture components share the same consistent set of univariate and pairwise marginals, it is natural to expect that this ratio does not deviate significantly from $1$, thereby leading to high-quality samples. Indeed, our empirical analysis demonstrates that the aforementioned intuition does hold. 

Note that we do not actually need to compute $P(\e)$ to obtain the sample weights when computing expectations. We can use the unnormalized weight $w'(y,\x \mid \e)=P(\e \mid y)$ as $P(\e)$ is a multiplicative constant given $\e$, thereby leading to a self-normalizing importance sampling algorithm. The expectation of any function $f(\X)$ over $P$ can be estimated using samples $\mathcal{D}=\{\boldsymbol{x}y[1], \ldots, \boldsymbol{x}y[M]\}$ from $Q$ as:
\begin{align*}
&\hat{\boldsymbol{E}}_{\mathcal{D}}(f)
=\frac{\sum_{m=1}^M f(\boldsymbol{x}[m]) w(\boldsymbol{x}y[m])}{\sum_{m=1}^M w(\boldsymbol{x}y[m])} \\
&\quad=\frac{\sum_{m=1}^M f(\boldsymbol{x}[m]) P(\e \mid y[m])}{\sum_{m=1}^M P(\e \mid y[m])} \\
&\quad=\frac{\sum_{m=1}^M f(\boldsymbol{x}[m]) w'(\boldsymbol{x}y[m])}{\sum_{m=1}^M w'(\boldsymbol{x}y[m])}
\end{align*}
\paragraph{Collapsed Sampling} Observe that the sample weights $w(y,\x \mid \e)=\frac{P(\e \mid y)}{P(\e)=w(y\mid \e)}$ only depend on $\e$ and $\y$ and are independent of $\x$.
Given an arbitrary function $f(\x)$, this allows to effectively ``push the expectation inside" to the tree level, and freely leverage any estimation method available for estimating the expectation of $f(\x)$ on a tree distribution. This amounts to a form of collapsed sampling~\citep{koller2009probabilistic}:

\begin{align*}
&\boldsymbol{E}_{\x,y \sim P(\cdot \mid \e)}(f(\x))=\sum_{\x y} P(y \mid \e)\cdot P(\x \mid y\e) \cdot f(\x)\\
&\quad=\sum_{\x y} w(y,\x \mid \e)\cdot P(y)\cdot P(\x \mid y\e) \cdot f(\x)\\
&\quad=\sum_{y} P(y) \sum_{\x} w(y,\x \mid \e) \cdot P(\x \mid y\e)\cdot f(\x)\\
&\quad=\sum_{y} P(y)\cdot w(y \mid \e) \sum_{\x} \left(P(\x \mid y\e)\cdot f(\x)\right)\\
&\quad=\sum_{y} P(y)\cdot w(y \mid \e) \cdot \boldsymbol{E}_{\x \sim P(\cdot \mid y\e)}f(\x)
\end{align*}

Our empirical estimator then becomes
\begin{align*}
\hat{\boldsymbol{E}}_{\mathcal{D}}(f)
=&\frac{\sum_{m=1}^M w'(y[m]) \boldsymbol{E}_{\x \sim (\cdot \mid y\e)}(f(\x)) }{\sum_{m=1}^M w'(y[m])}
\end{align*}

Intuitively, we sample a spanning tree, compute the desired quantity in the corresponding tree distribution, and weight the estimate appropriately. We are thus able to drastically speed up convergence by leveraging the whole suite of exact and approximate techniques available for estimation in tree distributions which have been extensively studied in the literature. For instance, as conditionals of the form $P(X_i=1 \mid \e)$ are tractable in tree distributions, we can efficiently estimate ${\Pr}_{\text{MoAT}}(X_i=1 \mid \e)$ as 
\begin{align*}
\hat{\Pr}_{\text{MoAT}}(X_i\!=\!1 \mid \e)
=&\frac{\sum_{m=1}^M w'(y[m]) P(X_i=1 \mid y[m]\e) }{\sum_{m=1}^M w'(y[m])}
\end{align*}

\paragraph{Empirical Evaluation}Empirically, we evaluate our importance sampling algorithm and the collapsed importance sampling algorithm against a standard Gibbs sampling algorithm~\citep{gelfand1990sampling}, which is enabled by tractable likelihood computation on the MoAT model. In our experiments, we focus on posterior marginal inference: we fix evidence $\e$ of various sizes, and estimate univariate marginals of the remaining variables conditioned on the evidence $P(X_i \mid \e)$. To illustrate speed of convergence to the true value, we require to exactly compute these ground-truth conditionals. To that end, we limit ourselves to a MoAT model on the 16 variable NLTCS dataset from the Twenty Datasets benchmark, where we can exactly compute MoAT marginals and conditionals by exhaustive enumeration. We use average KL-divergence as our metric to assess the speed of convergence:
\begin{align*}
    &D_{\mathrm{KL}}(P \| \hat{P}) \\
    &=\!\sum_{X_i \in \X \setminus E}\!P(x_i\!\mid\!\e) \log \frac{P(x_i\!\mid\!\e)}{\hat{P}(x_i\!\mid\!\e)}
    +P(\overline{x_i}\!\mid \e)\!\log \frac{P(\overline{x_i}\!\mid\!\e)}{\hat{P}(\overline{x_i}\!\mid\!\e)}
\end{align*}

\begin{figure}[thb]
    \centering
    \includegraphics[width=0.7\linewidth]{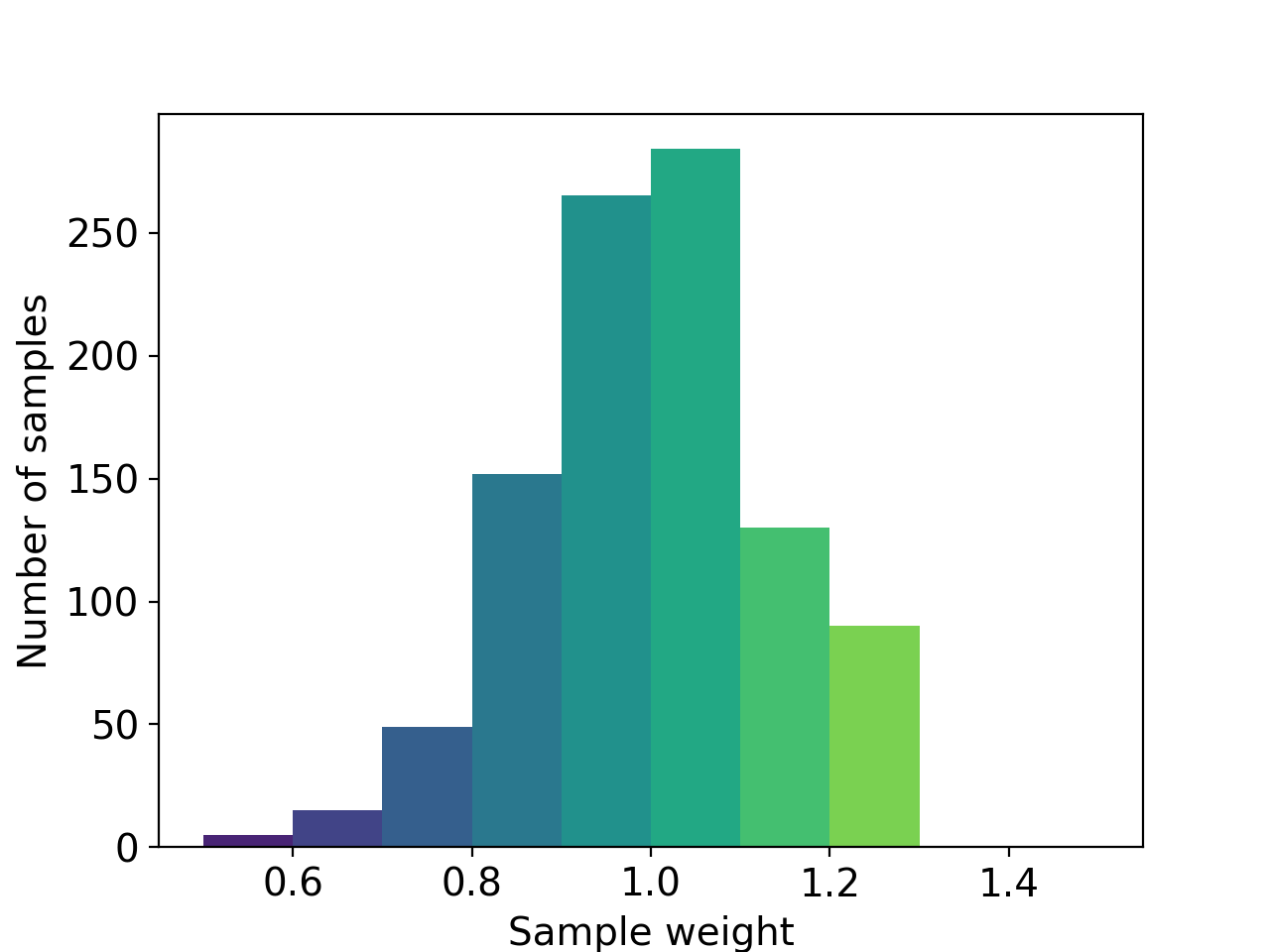}
    \caption{Distribution of sample weights for $\abs{\e}=4$.} 
    \label{fig:weights-hist}
\end{figure}

As we see \cref{fig:kld-convergence}, the importance sampling and collapsed importance sampling converge orders of magnitude faster than Gibbs sampling.
These results are all the more impressive when we account for the superior computational complexity of importance sampling. The bottleneck in the importance sampling algorithm is the spanning tree sampling, leading to a time complexity of $\mathcal{O}(n^\omega)$. However, each sample in Gibbs sampling requires $n$ likelihood estimation queries, resulting in a complexity of $\mathcal{O}(n \cdot n^\omega)$. Further, we observe that the importance sampling algorithm produces very high quality samples as illustrated by the closeness of sample weights to $1$ (\cref{fig:weights-hist}).

\section{CONCLUSION}
In this paper, we propose a novel class of generative models called mixture of all trees (MoAT), which strikes a new balance between expressivity and tractability. Despite being a mixture over super-exponentially many tree-shaped distributions, we show that 
it allows for tractable computation of~(normalized) likelihood. Besides, learning a MoAT model does not involve the problem of structure learning, which plagues most probabilistic graphical models. 

While we prove hardness of certain classes of queries such as MAP, we demonstrate how MoAT's foundation in tree-shaped models allows us to naturally obtain extremely fast approximate inference algorithms by interpreting it as latent variable model with clear semantics and leveraging tractability of its underlying mixture components. Empirically, we see that MoAT achieves state-of-the-art performance on a variety of density estimation tasks, outperforming powerful probabilistic models such as HCLTs. We leave it to future work to explore MoAT's potential to scale to non-tabular data such as images and text.

We hope that MoAT opens up interesting questions that push the boundaries of tractability and expressive power for probabilistic graphical models.

\section*{Acknowledgements}
We thank the reviewers for their thoughtful feedback towards improving this paper. 
This work was funded in part by the DARPA Perceptually-enabled Task Guidance (PTG) Program under
contract number HR00112220005, and NSF grants \#IIS-1943641, \#IIS-1956441, and \#CCF-1837129.

\bibliography{main}

\onecolumn
\appendix
\section{Complete Proofs}
\setcounter{lem}{0}
This is section, we present lemmas whose proofs were deferred to the appendix.
\lem{For any distribution $\Pr(\cdot)$ over binary random variables $X_1, \dots, X_n$, there exists a set of parameters~(i.e., $p_v$ and $p_{uv}$) in our hypothesis space such that $\Pr(X_u) = P_{u}(X_u)$ and $\Pr(X_u, X_v) = P_{uv}(X_u, X_v)$ for all $1 \leq u, v \leq n$; i.e., the univariate and pair-wise marginals of $\Pr$ are the same as  $P_{u}$ and and $P_{uv}$.}
\begin{proof}~ Pick MoAT parameters $p_v=\Pr(X_v\!=1)$ and $p_{uv}=\Pr(X_u\!=\!1, X_v\!=\!1)$. By construction, the summation constraints are satisifed. Thus, it suffices to check that all the univariate and pairwise marginals are non-negative. For any $v \in V$, we have that $P(X_v=1)=p_v \in [0,1]$. Then $P(X_v=0)=1-p_v \in [0,1]$ as desired. Further, for every $\{u,v\} \in {V \choose 2}$, $P(X_u=1, X_v=0)=p_u-p_{uv},P(X_u=0, X_v=1)=p_v-p_{uv},\text{ and } P(X_u=0, X_v=0)=1-p_{uv}-(p_u-p_{uv})-(p_v-p_{uv})=p_{uv}-(p_u+p_v-1) \geq 0$ since $P(X_u=1, X_v=1) \in [max(0,p_u+p_v-1),min(p_u,p_v)]$ Hence, the univariate and pair-wise marginals of $\Pr$ are the same as  $P_{u}$ and and $P_{uv}$, as desired.

    
.
\end{proof}

\lem{Define $\mathsf{SST}$ as the following decision problem: given a connected graph $G=(V,E)$ and a subset $K \subset V$ of the vertices, decide if there exists a spanning tree of $G$ whose leaves are exactly $K$. $\mathsf{SST}$ is NP-hard.} 
\begin{proof}
    We proceed via reduction from $\mathsf{HAMILTONIAN-PATH}$. Observe that a spanning tree with exactly two leaves is a Hamiltonian path between the two leaves. Given $G=(V,E)$, we iterate over all pairs of vertices $\{i,j\}$, and query the  $\mathsf{SST}$ oracle for the existence of spanning tree with $K=\{i,j\}$. Then, $G$ has a Hamiltonian path if and only if there exists at least one pair of vertices for which the decision of the $\mathsf{SST}$ oracle is $\mathsf{YES}$.
\end{proof}

\lem{Let $\x$ be a complete assignment, and denote by $\mathsf{ONES}(\x)$ the set of variables are are set to $1$ in $\x$. Denote by $\abs{\x}$ the value $\abs{\mathsf{ONES}(\x)}$. Denote by $\mathsf{LEAVES}(T)$ the set of leaves of a spanning tree $T$. Let $k$ be the number of spanning trees $T$ of $G$ with $\mathsf{ONES}(\x) \subseteq \mathsf{LEAVES} (T)$. Then, 
$\begin{cases} 
        \frac{k}{\epsilon^{n-2}} \leq Z \cdot p(\x) \leq \frac{k}{\epsilon^{n-2}} +\frac{n^{n-2}}{\epsilon^{n-3}}, &  \abs{\x}\%2=0\\
        \frac{-k}{\epsilon^{n-2}}+\frac{-n^{n-2}}{\epsilon^{n-3}} \leq Z \cdot p(\x) \leq \frac{-k}{\epsilon^{n-2}} , & \abs{\x}|\%2=1\\
   \end{cases}$

}
\begin{proof}
We will compute the values of the MOAT likelihood function $p$ for any complete assignment $\x$. 
\begin{itemize}
    \item Case 1: $\mathsf{ONES}(\x) \subseteq \mathsf{LEAVES} (T)$ 
    \begin{align*}
    &\left(\prod_{v \in V} P_v\left(x_v\right)\right) \left(\prod_{(u,v) \in E} w_{(u,v)} \cdot \frac{P_{uv}(x_u,x_v)}{P_u(x_u) \cdot P_v(x_v)} \right)\\
    =&\abs{\left(\prod_{v \in V} P_v\left(x_v\right)\right) \left(\prod_{(u,v) \in E} w_{(u,v)} \cdot \frac{P_{uv}(x_u,x_v)}{P_u(x_u) \cdot P_v(x_v)} \right)} \cdot (-1)^{\abs{\x}} \comment{Since $\frac{P_{uv}}{P_u \cdot P_v} \geq 0$ and $P_v(x_v) < 0 \iff x_v=1$ }\\ 
    =& \abs{\frac{\prod_{(u,v) \in E}w_{(u,v)} P_{uv}\left(x_u, x_v\right)}{\prod_{v \in V} P_v\left(x_v\right)^{\operatorname{deg} v-1}}}\cdot (-1)^{\abs{\x}}\\
    =& \abs{\frac{\prod_{(u,v) \in E} \epsilon}{\prod_{v \in V} \epsilon^{\operatorname{deg} v-1}}}\cdot (-1)^{\abs{\x}}\\
    =& \abs{\frac{\prod_{(u,v) \in E} \epsilon}{\prod_{v \in V} \epsilon^{\operatorname{deg} v-1}}}\cdot (-1)^{\abs{\x}}\\
    =& \abs{\frac{\epsilon^{n-1}}{ \epsilon^{2n-3}}}\cdot (-1)^{\abs{\x}}\\
    =& \frac{1}{\epsilon^{n-2}}\cdot (-1)^{\abs{\x}}
    \end{align*}
    \item Case 2: $\mathsf{ONES}(\x) \not\subseteq \mathsf{LEAVES} (T)$\\
    In this case $\gamma \geq 1$ internal nodes (nodes with degree more than one) are assigned a value of 1. Then similarly, 
    \begin{itemize}
        \item If $\abs{\x}\%2=0$, we obtain that 
         \begin{align*}
        0 \leq & \left(\prod_{v \in V} P_v\left(x_v\right)\right) \left(\prod_{(u,v) \in E} w_{(u,v)} \cdot \frac{P_{uv}(x_u,x_v)}{P_u(x_u) \cdot P_v(x_v)} \right)\\
        =&\abs{\left(\prod_{v \in V} P_v\left(x_v\right)\right) \left(\prod_{(u,v) \in E} w_{(u,v)} \cdot \frac{P_{uv}(x_u,x_v)}{P_u(x_u) \cdot P_v(x_v)} \right)}  \\ 
        =& \abs{\frac{\prod_{(u,v) \in E} \epsilon}{\prod_{v \in V} P_v\left(x_v\right)^{\operatorname{deg} v-1}}}\\
        \leq& \abs{\frac{\epsilon^{n-1}}{ \epsilon^{2n-4}}}\\
        \leq& \frac{1}{\epsilon^{n-3}}
        \end{align*}
        \item If $\abs{\x}\%2=1$, we obtain that 
         \begin{align*}
        0 \geq & \left(\prod_{v \in V} P_v\left(x_v\right)\right) \left(\prod_{(u,v) \in E} w_{(u,v)} \cdot \frac{P_{uv}(x_u,x_v)}{P_u(x_u) \cdot P_v(x_v)} \right)\\
        =&-\abs{\left(\prod_{v \in V} P_v\left(x_v\right)\right) \left(\prod_{(u,v) \in E} w_{(u,v)} \cdot \frac{P_{uv}(x_u,x_v)}{P_u(x_u) \cdot P_v(x_v)} \right)}  \\ 
        =& -\abs{\frac{\prod_{(u,v) \in E} \epsilon}{\prod_{v \in V} P_v\left(x_v\right)^{\operatorname{deg} v-1}}}\\
        \geq& -\abs{\frac{\epsilon^{n-1}}{ \epsilon^{2n-4}}}\\
        \geq& -\frac{1}{\epsilon^{n-3}}
        \end{align*}
    \end{itemize}
\end{itemize}
    
As the maximum number of spanning trees on a graph with $n$ vertices is $n^{n-2}$, we obtain the desired bounds:

\begin{itemize}
        \item If $\abs{\x}\%2=0$, we obtain that 
         \begin{align*}
         Z \cdot p(\x)=& \sum_{T \in \mathsf{ST}(G)} \left(\prod_{v \in V} P_v\left(x_v\right)\right) \left(\prod_{(u,v) \in E} w_{(u,v)} \cdot \frac{P_{uv}(x_u,x_v)}{P_u(x_u) \cdot P_v(x_v)} \right) \\
         =& \sum_{\substack{T \in \mathsf{ST}(G) \\  K \subseteq \mathsf{LEAVES}(T)}} \left(\prod_{v \in V} P_v\left(x_v\right)\right) \left(\prod_{(u,v) \in E} w_{(u,v)} \cdot \frac{P_{uv}(x_u,x_v)}{P_u(x_u) \cdot P_v(x_v)} \right) \\
         &+ \sum_{\substack{T \in \mathsf{ST}(G) \\K \not\subseteq \mathsf{LEAVES}(T)}} \left(\prod_{v \in V} P_v\left(x_v\right)\right) \left(\prod_{(u,v) \in E} w_{(u,v)} \cdot \frac{P_{uv}(x_u,x_v)}{P_u(x_u) \cdot P_v(x_v)} \right) \\
         \geq& \sum_{\substack{T \in \mathsf{ST}(G) \\ K \subseteq \mathsf{LEAVES}(T)}} \frac{1}{\epsilon^{n-2}} + \sum_{\substack{T \in \mathsf{ST}(G) \\ K \not\subseteq \mathsf{LEAVES}(T)}} 0\\
         \geq & \frac{k}{\epsilon^{n-2}} \\
         \text{Similarly }Z \cdot p(\x)\leq&  \sum_{\substack{T \in \mathsf{ST}(G) \\  K \subseteq \mathsf{LEAVES}(T)}} \frac{1}{\epsilon^{n-2}} + \sum_{\substack{T \in \mathsf{ST}(G) \\ K \not\subseteq \mathsf{LEAVES}(T)}} \frac{1}{\epsilon^{n-3}}\\
         \leq & \frac{k}{\epsilon^{n-2}} + \frac{n^{n-2}}{\epsilon^{n-3}} \\
        \end{align*}
        
        \item If $\abs{\x}\%2=1$, we similarly obtain that 
        \begin{align*}
        Z \cdot p(\x)\leq & \frac{-k}{\epsilon^{n-2}} \\
        Z \cdot p(\x)\geq & \frac{-k}{\epsilon^{n-2}} + \frac{-n^{n-2}}{\epsilon^{n-3}} \\
        \end{align*}
        
\end{itemize}

Thus, $\begin{cases} 
        \frac{k}{\epsilon^{n-2}} \leq Z \cdot p(\x) \leq \frac{k}{\epsilon^{n-2}} +\frac{n^{n-2}}{\epsilon^{n-3}}, &  \abs{\x}\%2=0\\
        \frac{-k}{\epsilon^{n-2}}+\frac{-n^{n-2}}{\epsilon^{n-3}} \leq Z \cdot p(\x) \leq \frac{-k}{\epsilon^{n-2}} , & \abs{\x}|\%2=1\\
   \end{cases}$
   as desired.

\end{proof}

\lem{The number of spanning trees $T$ with $K = \mathsf{LEAVES}(T)$ is given by $\abs{\lfloor Z \epsilon^{n-2} f(\e) \rceil}$. } 

\begin{proof}

Since the number of spanning trees $T$ with $K \subseteq \mathsf{LEAVES}(T)$ is given by $\abs{\lfloor Z \cdot \epsilon^{n-2} \cdot p(\x) \rceil}$, from the inclusion-exclusion formula we obtain that the number of spanning trees $T$ with $K = \mathsf{LEAVES}(T)$ (upto sign) is given by
\begin{align*}
    & \sum_{K\subseteq L} (-1)^{\abs{L}}\sum_{T \in \mathsf{ST}(G)} \mathds{1}(L \subseteq \mathsf{LEAVES}(T)) \\
    =&\sum_{val(z_1)} \sum_{val(z_2)} \ldots \sum_{val(z_k)} (-1)^{\abs{\x}}\abs{\lfloor Z \epsilon^{n-2} p(\x) \rceil} \\
    =& \sum_{val(z_1)} \sum_{val(z_2)} \ldots \sum_{val(z_k)} \lfloor Z \epsilon^{n-2} p(\x) \rceil \\ 
\end{align*}

Let $\{\x\}=\abs{\x - \lfloor \x \rceil}$. Since the $\{ Z \cdot \epsilon^{n-2} \cdot p(\x)\} 
\leq \epsilon \cdot n^{n-2}$ for all $\x$, we obtain that
\begin{align*}
     &\sum_{val(z_1)} \sum_{val(z_2)} \ldots \sum_{val(z_k)} \{ Z \cdot \epsilon^{n-2} \cdot p(\x)\}\\
     \leq & 2^{n} \cdot \epsilon \cdot n^{n-2} \\
     < & 2^{n} \cdot \frac{1}{2^{n+1}\cdot n^{n-2}} \cdot n^{n-2} \\
     < & \frac{1}{2}
\end{align*}

Thus, we obtain that the number of spanning trees $T$ with $K = \mathsf{LEAVES}(T)$ is given by 
\begin{align*}
    & \sum_{val(z_1)} \sum_{val(z_2)} \ldots \sum_{val(z_k)} \lfloor Z \cdot \epsilon^{n-2} \cdot p(\x) \rceil \\ 
    = & \lfloor \sum_{val(z_1)} \sum_{val(z_2)} \ldots \sum_{val(z_k)}  Z \cdot \epsilon^{n-2} \cdot p(\x) \rceil \\ 
    = & \lfloor Z \cdot \epsilon^{n-2} \cdot \sum_{val(z_1)} \sum_{val(z_2)} \ldots \sum_{val(z_k)}   p(\x) \rceil \\ 
    = & \lfloor Z \cdot \epsilon^{n-2} \cdot f(\e) \rceil \\ 
\end{align*}

\end{proof}

\section{Parameterization for Categorical Variables}
Consider the MoAT pairwise marginal matrix $P_{uv}^{(k_u\times k_v)}$ defined by $P_{uv}[i][j]=P(X_u=i, X_v=j)$. 

\subsection{Relation to Doubly Stochastic Matrices}
This MoAT pairwise marginal matrix is closely related to the class of matrices called doubly stochastic matrices, where all the entries are between $0$ and $1$ and the rows and columns sum to $1$. Recall that for the MoAT pairwise marginal matrix, we similarly require that the rows and columns sum to the corresponding univariate marginals. The set of $k \times k$ doubly stochastic matrices (often referred to as the Birkhoff polytope) lies in a $(k-1)\times(k-1)$ affine subspace of $\mathbb{R}^{k\times k}$, and to the best of our knowledge there is no known valid and fully general parameterization for this class of matrices that allows for unconstrained optimization. Similarly, the pairwise marginal matrices which are uniquely defined by the values $P_{uv}[i][j]$ for $(i,j) \in \{1,2,\cdots , k-1\} \times \{1,2,\cdots , k-1\}$, also lie in a $(k-1)\times(k-1)$ affine subspace of $\mathbb{R}^{k\times k}$, and there is unfortunately, to the best of our knowledge, no known valid and fully general parameterization for this class of matrices that allows for unconstrained optimization.

However, we instead propose a valid parameterization of MoAT pairwise marginal matrices that is not fully general, but has $\min(k_u,k_v)-1$ free parameters (as opposed to a fully general paramterization with $(k_u-1)\times (k_v-1)$ parameters) that can be learnt in an unconstrained manner.

\subsection{Proposed MoAT Parameterization}

First we consider the case of square pairwise marginal matrices $P_{uv}^{(k \times k)}$. We define it inductively, defining a parameterization for the first $l \times l$ submatrix of $P_{uv}$ (denoted $P_l$)for $l \in [2,k]$. One can interpret this a defining a parameterization for the marginal distribution $P(X_u \in [1,l], X_v \in [1,l])$, in a way that preserves the relative proportion of univariate marginals. 

\begin{itemize}
    \item \textbf{Base case ($l=2$):}\\ This is identical to having binary random variables $X_u$ and $X_v$ with the following univariate marginals. \begin{itemize}
        \item $P_2(X_u=1)=\frac{P(X_u=1)}{P(X_u=1)+P(X_u=2)}$
        \item $P_2(X_u=2)=\frac{P(X_u=2)}{P(X_u=1)+P(X_u=2)}$
        \item $P_2(X_v=1)=\frac{P(X_v=1)}{P(X_v=1)+P(X_v=2)}$
        \item $P_2(X_v=2)=\frac{P(X_v=2)}{P(X_v=1)+P(X_v=2)}$
        
    \end{itemize}
    The matrix can be parameterized by a single parameter $\lambda_2=P_2(X_u=1, X_v=1)$ as shown in Lemma 1. 
    \item \textbf{Inductive case ($l>=3$):}\\ Assume we have have a parameterization for first $(l-1) \times (l-1)$ submatrix $P_{l-1}$ of $P_{uv}$. Pick $\lambda_l  \in [max(0,\frac{\sum_{t=1}^{l-1}P(X_u=t)}{\sum_{t=1}^{l}P(X_u=t)}+\frac{\sum_{t=1}^{l-1}P(X_v=t)}{\sum_{t=1}^{l}P(X_v=t)}-1),min(\frac{\sum_{t=1}^{l-1}P(X_u=t)}{\sum_{t=1}^{l}P(X_u=t)},\frac{\sum_{t=1}^{l-1}P(X_v=t)}{\sum_{t=1}^{l}P(X_v=t)})]$. 
    
    Then, define $P_l$ as follows: \\
    $P_l[i][j]=\begin{cases} 
        \lambda_l \times P_{l-1}[i][j], & i<l, j<l\\
        \frac{P(X_u=i)}{\sum_{t=1}^{l}P(X_u=t)} - \lambda_l \times \frac{P(X_u=i)}{\sum_{t=1}^{l-1}P(X_u=t)}, & i<l, j=l\\
        \frac{P(X_v=j)}{\sum_{t=1}^{l}P(X_v=t)} - \lambda_l \times \frac{P(X_v=j)}{\sum_{t=1}^{l-1}P(X_v=t)}, & i=l, j<l\\
        1- \frac{\sum_{t=1}^{l-1}P(X_u=t)}{\sum_{t=1}^{l}P(X_u=t)} - \frac{\sum_{t=1}^{l-1}P(X_v=t)}{\sum_{t=1}^{l}P(X_v=t)} + \lambda_l, & i=l, j=l\\
   \end{cases}$

   By choice of $\lambda_l$, all the entries of this matrix are non-negative. It now suffices to check that the univariate marginals are in proportion. 
   \begin{itemize}
       \item For $i<l$,
       \begin{align*}
           P_l(X_u=i) &= \sum_{j=1}^l P_l[i][j] \\
            &= \sum_{j=1}^{l-1} P_l[i][j] + P_l[i][l] \\
            &= \sum_{j=1}^{l-1} \left(\lambda_l \times P_{l-1}[i][j]\right) + \frac{P(X_u=i)}{\sum_{t=1}^{l}P(X_u=t)} - \lambda_l \times \frac{P(X_u=i)}{\sum_{t=1}^{l-1}P(X_u=t)} \\
            &= \frac{P(X_u=i)}{\sum_{t=1}^{l}P(X_u=t)}
       \end{align*}
       as desired.\\

       \item For $i=l$,
       \begin{align*}
           P_l(X_u=l) =& \sum_{j=1}^l P_l[l][j] \\
            =& \sum_{j=1}^{l-1} P_l[l][j] + P_l[l][l] \\
            =& \sum_{j=1}^{l-1} \left(\frac{P(X_v=j)}{\sum_{t=1}^{l}P(X_v=t)} - \lambda_l \times \frac{P(X_v=j)}{\sum_{t=1}^{l-1}P(X_v=t)} \right) \\ &+ 1- \frac{\sum_{t=1}^{l-1}P(X_u=t)}{\sum_{t=1}^{l}P(X_u=t)} - \frac{\sum_{t=1}^{l-1}P(X_v=t)}{\sum_{t=1}^{l}P(X_v=t)} + \lambda_l \\
            =& \sum_{j=1}^{l-1} \left(\frac{P(X_v=j)}{\sum_{t=1}^{l}P(X_v=t)}  \right) + 1- \frac{\sum_{t=1}^{l-1}P(X_u=t)}{\sum_{t=1}^{l}P(X_u=t)} - \frac{\sum_{t=1}^{l-1}P(X_v=t)}{\sum_{t=1}^{l}P(X_v=t)} \\
            =& 1- \frac{\sum_{t=1}^{l-1}P(X_u=t)}{\sum_{t=1}^{l}P(X_u=t)} \\
            =& \frac{P(X_u=l)}{\sum_{t=1}^{l}P(X_u=t)} 
       \end{align*}
        as desired.\\

        \item By symmetry, the desired results hold for $P_l(X_v=j)$ for all $ 1 \leq j \leq l$.
   \end{itemize}
\end{itemize}

Observe that since $\sum_{t=1}^{k}P(X_u=t) = 1$, $P_k(X_u=t)=P(X_u=t)$ for all $ 1 \leq t \leq k$. Similarly, $P_k(X_v=t)=P(X_v=t)$ for all $ 1 \leq t \leq k$. Thus, $P_k$ is the desired $k \times k$ MoAT pairwise marginal matrix, with learnable parameters $\lambda_2 \cdots \lambda_k$. 

Lastly, observe that this parameterization generalizes to non-square matrices too. Without loss in generality, assume $P$ is $k_u \times k_v$ with $k_u < k_v$. First, we can parameterize the first $2 \times k_v - k_u - 2$ submatrix by a single parameter. Then, we can add $k_u-2$ scaling parameters $\lambda_i$ as in the case of the square matrix to obtain a parameterization for the whole matrix. Note that the total number of free parameters in this parameterization is $\min(k_u,k_v)-1$.

\section{Experimental Setup}
All experiments were performed on Intel(R) Xeon(R) Gold 5220 CPU @ 2.20GHz. For the experiments on the Twenty Dataset density estimation benchmark, the MoAT model is trained with two sets of hyperparameters: (1) for the datasets with $< 500$ random variables, the model is trained with batch\_size = 1024 and learning rate = 0.05 and (2) for the datasets with $\geq 500$ random variables, the model is trained with batch\_size = 64 and learning rate = 0.01. All models are trained for 50 epochs with early stopping: the test log-likelihood corresponding to the epoch with the best validation log-likelihood is reported. The total training time for all datasets takes roughly a day on one NVIDIA RTX A5000 gpu. Complete code and datasets for all the experiments can be found at \texttt{https://github.com/UCLA-StarAI/MoAT}.

\vfill

\end{document}